\documentclass[journal]{IEEEtran}
\usepackage{amsmath}
\usepackage[ruled,vlined]{algorithm2e}
\SetKwInput{KwReq}{Requires}
\usepackage{algpseudocode}
\usepackage{tikz}
\usepackage{graphicx}
\usepackage{url}
\usepackage{caption}
\usepackage{booktabs}
\usepackage{newunicodechar}
\newunicodechar{₄}{$_4$}
\title{\textbf{Detection of Intoxicated Individuals from Facial Video Sequences via a Recurrent Fusion Model}}
\title{Detection of Intoxicated Individuals from Facial Video Sequences via a Recurrent Fusion Model}

\author{
    Bita~Baroutian,
    Atefe~Aghaei,
    Mohsen~Ebrahimi~Moghaddam\thanks{Corresponding author: m\_moghadam@sbu.ac.ir}%
    \\
    Shahid Beheshti University, Tehran, Iran \\
    bi.baroutian@mail.sbu.ac.ir, a\_aghaei@sbu.ac.ir,\ m\_moghadam@sbu.ac.ir
}

\begin{document}

\maketitle

\begin{abstract}
Alcohol consumption is a significant public health concern and a major cause of accidents and fatalities worldwide. This study introduces a novel video-based facial sequence analysis approach dedicated to the detection of alcohol intoxication.  The method integrates facial landmark analysis via a Graph Attention Network (GAT) with spatiotemporal visual features extracted using a 3D ResNet. These features are dynamically fused with adaptive prioritization to enhance classification performance. Additionally, we introduce a curated dataset comprising 3,542 video segments derived from 202 individuals to support training and evaluation. Our model is compared against two baselines: a custom 3D-CNN and a VGGFace+LSTM architecture. Experimental results show that our approach achieves 95.82\% accuracy, 0.977 precision, and 0.97 recall, outperforming prior methods. The findings demonstrate the model’s potential for practical deployment in public safety systems for non-invasive, reliable alcohol intoxication detection. 

\end{abstract}

\section{Introduction}
Alcohol consumption remains a significant public safety challenge, particularly when it negatively affects cognitive functions, physical coordination, and judgment. Such impairments substantially increase the likelihood of hazardous behaviors, notably driving under the influence (DUI), which continues to be a persistent global issue despite stringent legal measures and public awareness campaigns.\cite{Alonso2015} Governments and regulatory bodies worldwide have implemented a range of interventions, including stricter penalties, license suspensions, and advanced roadside testing protocols, to mitigate the risks associated with alcohol-impaired driving. However, the efficacy of these measures remains limited, as traditional alcohol detection methods, predominantly breath analyzers, present notable challenges.\cite{Olson2025} These devices, while effective in controlled settings, often suffer from drawbacks such as high operational costs, hygiene concerns, and inaccuracies stemming from residual moisture after prolonged use \cite{chang2021drunk}. Consequently, there is an increasing need for innovative, non-intrusive, and accurate methods to detect alcohol impairment, particularly in dynamic real-world environments.

Research into alcohol intoxication detection methods has gained considerable momentum in recent years, driven by advancements in machine learning and computer vision. Traditional approaches, primarily based on breath analysis, are now being complemented and, in some cases, replaced by more sophisticated techniques\cite{Jones2020}. These novel methods can be categorized into four main domains: thermal infrared imaging, facial image analysis, gait pattern analysis, and video-based facial sequence analysis. Thermal infrared imaging capitalizes on detecting temperature changes induced by alcohol consumption, as vasodilation resulting from intoxication can elevate skin surface temperature \cite{menon2019driver}. Gait pattern analysis, in contrast, assesses movement characteristics such as stride length, step width, and balance irregularities, which are indicative of compromised movement control due to alcohol intake \cite{park2023deep}. Additionally, video-based facial analysis, a recent development in the field, leverages deep learning algorithms to remotely detect facial cues associated with intoxication, thereby offering a non-intrusive alternative to traditional methods. Unlike breath analyzers that necessitate direct interaction, video-based approaches can operate passively, making them particularly suitable for public safety monitoring and workplace compliance applications \cite{mehta2019dif}.

Despite the progress made through these innovative methods, practical implementation challenges persist. Traditional breath analyzers, while reliable under specific conditions, are often intrusive, raising concerns about user comfort and compliance. Thermal imaging methods are sensitive to ambient temperature fluctuations\cite{koukiou2022thermal}, and gait analysis can yield inconsistent results due to individual differences in walking styles\cite{park2023deep}. Addressing these challenges is essential for the development of scalable and practical drunk person detection systems, capable of functioning effectively in varied real-world scenarios.

The key contributions of this paper are summarized as follows:

1. We propose a novel video-based facial sequence analysis framework to detect alcohol consumption. The approach leverages advanced deep learning techniques to analyze facial landmarks and spatiotemporal features, providing a non-intrusive, accurate, and remote detection solution.

 2. The proposed model integrates facial landmark analysis through a Graph Attention Network (GAT) with visual spatiotemporal feature extraction using a 3D ResNet. This dual-branch architecture captures both fine-grained facial dynamics and broader motion cues, enhancing detection robustness.

 3. A key innovation of our method is a weighted fusion strategy that dynamically balances contributions from spatial and temporal features. 

The rest of this paper is organized as follows: Section 2 reviews related work on alcohol intoxication detection across various modalities. Section 3 introduces the proposed recurrent fusion framework, detailing its architectural components and adaptive fusion strategy. Section 4 presents the curated dataset, experimental setup, and implementation details. Section 5 reports the results of ablation studies and comparative evaluations against baseline models. Finally, Section 6 concludes the paper and outlines potential directions for future research.

\section{Related Works}
The detection of alcohol consumption has remained a longstanding research challenge, evolving significantly over the last century at the nexus of physiology, computer vision, and mobile sensing. Contemporary methodologies fall into two principal categories: (1) direct physiological measurements and (2) machine learning-based approaches that utilize behavioral or appearance cues. 

\subsection{Physiological Measurement Techniques}
 Breath analysis, regarded as the clinical and legal standard, quantifies blood alcohol concentration (BAC) via exhaled ethanol. Initial devices such as Harger’s Drunkometer relied on chemical colorimetric reactions with acidic KMnO₄ solutions\cite{mcvean2019breathalyzer}, while Borkenstein’s Breathalyzer (1954) introduced photometric analysis for increased portability and efficiency  \cite{borkenstein1961breathalyzer}. Modern breath analyzers utilize either infrared (IR) spectroscopy to capture ethanol’s specific absorption bands \cite{berger2002breath} or fuel-cell sensors for real-time current generation via ethanol oxidation  through ethanol oxidation \cite{ozoemena2018fuel}. However, these systems suffer from practical constraints: calibration drift, environmental susceptibility (e.g., humidity, volatile compounds), hygiene concerns, and the inability to support continuous or large-scale deployment  \cite{webmd2023breath}. 
\subsection{Machine Learning and Vision-Based Techniques}
In response to the limitations of physiological techniques, machine learning approaches have emerged, emphasizing non-invasive detection strategies. These include: 

\textbf{Thermal Imaging:} exploits the alcohol-induced vasodilation effects of ethanol, where intoxication typically lowers forehead temperature while increasing nasal temperature. Kokyo et al.\cite{koukiou2022thermal} demonstrated that segmenting iso-thermal regions in multi-frame thermal images and utilizing morphological descriptors within Support Vector Machines (SVMs) could achieve a mean accuracy of 76\% on a dataset of 41 subjects, increasing to 86\% when focusing on forehead-specific regions. While contact-free, this approach faces challenges related to ambient temperature sensitivity, weak thermal contrast under physiological stress or illness, and the limitations posed by the high cost and resolution of long-wave IR cameras.

\textbf{Gait analysis:} another promising area for deep learning-based alcohol detection, capitalizes on the impairments in human walking stability induced by alcohol consumption. Park et al.\cite{park2023deep} developed a method generating Gait Energy Images (GEIs) from MOG2-segmented silhouettes, training a Convolutional Neural Network (CNN) to achieve 74.9\% accuracy when detecting intoxication in participants wearing “Drunk-Busters” goggles. In mobile sensing contexts, Li et al. \cite{li2021bac} transformed six-axis inertial data into Gramian Angular Fields for shallow CNN input and employed a Bi-LSTM with soft attention to estimate BAC directly. Despite achieving the lowest Root Mean Square Error (RMSE), these models required extensive data preprocessing and remained susceptible to variability in gait patterns, footwear changes, and smartphone positioning.

\textbf{Facial Image Analysis: }Static image-based methods utilize visual cues such as redness, facial asymmetry, or landmark displacement.  Chang et al. \cite{chang2021drunk} proposed a deep learning pipeline that combined a lightweight VGG model for age classification with a DenseNet for intoxication detection, achieving up to 94\% accuracy within specific age groups. However, generalization across diverse age groups significantly declined, with accuracy dropping to 42–45\%, indicating demographic bias .Takahashi et al. \cite{takahashi2015drinking} demonstrated that RGB differences from cheek patches, processed through a Multi-Layer Perceptron (MLP), could reach 79\% accuracy by detecting alcohol-induced erythema. Roberts et al.\cite{willoughby2019drunkselfie} employed a combination of 68 facial landmarks, forehead redness, edges, and wrinkle cues, using a Random Forest classifier to achieve 81\% accuracy on the studio-quality “3 Glasses Later” dataset . Additionally, Pai et al. \cite{kamath2021graph} enhanced detection by modeling the landmark graph using a two-layer Graph Convolutional Network (GCN), attaining an accuracy of 88\% on approximately 200,000 augmented samples.  Although deep learning techniques have brought significant improvements in this area, challenges such as sensitivity to makeup, variations in lighting conditions, and reliance on synthetically augmented data remain unresolved. 

\textbf{Video and Multimodal Analysis:} Temporal models that integrate visual and auditory features demonstrate increased robustness.  Dahl et al. \cite{mehta2019dif} compiled the DIF YouTube corpus, featuring both sober and intoxicated video samples, and compared models including a VGG-Face CNN followed by LSTMs and parameter-efficient 3D-CNN blocks. The optimal visual pipeline achieved over 80\% accuracy with a 40\% reduction in parameters compared to a conventional 3D-ResNet. Notably, audio analysis, leveraging deep learning methods like LSTMs to process features such as pitch, loudness, intensity, and Mel-Frequency Cepstral Coefficients (MFCC), outperformed purely visual models. The integration of visual and audio streams yielded an accuracy of 88.4\%. Despite these advances, the scarcity of well-annotated multimodal datasets aligned for intoxication detection remains a substantial obstacle.

\textbf{Speech-Based Intoxication Detection:} 
Recent work emphasizes speech as a low-cost, ubiquitous modality. Amato et al. introduced a domain-adversarial neural network trained on the Alcohol Language Corpus (ALC), incorporating features from OpenSmile and Praat, achieving a balanced accuracy of 70.9\% and demonstrating generalizability across speakers by minimizing inter-speaker variability.\cite{amato2024beyond}.Similarly, Albuquerque et al. employed Wav2Vec 2.0, a self-supervised, Transformer-based model, fine-tuned on ALC, reaching a state-of-the-art unweighted average recall (UAR) of 73.3\%, outperforming earlier approaches such as ResNet and BiRNN-based models \cite{albuquerque2025intoxication}.This work highlighted the advantages of self-supervised pretraining and robust contextual embeddings in intoxication detection, even in the presence of inter-speaker and class imbalance challenges. 

Despite notable progress, prior research often prioritizes performance under controlled conditions or within limited demographic scopes, undermining real-world applicability. Therefore, this study aims to advance a model that combines high accuracy with robust generalization under unconstrained, real-world settings, thereby bridging the gap between laboratory efficacy and practical deployment. 

\section{Methodology}
This section presents the proposed framework for detecting alcohol intoxication through facial video sequence analysis. The framework integrates facial landmarks and visual spatiotemporal features using a dual-branch architecture with an adaptive fusion mechanism, aiming to enhance classification accuracy and computational efficiency in real-world conditions. 

\subsection{Overview of the framework}
As illustrated in Figure~1, the architecture of the proposed model is composed of several key modules: shot change detection, landmark extraction with graph modeling, 3D-ResNet-based visual feature extraction, LSTM-driven temporal modeling, and an adaptive fusion component for classification. This multi-stage pipeline is designed to efficiently extract relevant behavioral information while minimizing noise and redundancy. The full process is summarized in Algorithm~\ref{alg:fusion}.
\begin{algorithm}[htbp]
\caption{Intoxication Detection via Shot-Based Recurrent Fusion}
\label{alg:fusion}
\DontPrintSemicolon

\KwIn{Input video $V$}
\KwOut{Predicted class $\hat{y} \in \{\text{sober},\text{intoxicated}\}$}

\KwReq{
Learnable fusion parameter $\alpha \in [0,1]$
}

\tcp{Shot change detection and face-based shot selection}

$\mathcal{S} = \{s_1,\dots,s_K\} \leftarrow \text{TransNetV2}(V)$ \tcp*{shot boundaries}
$\mathcal{S}_f \leftarrow \emptyset$ \tcp*{shots containing at least one face}

\ForEach{shot $s_k \in \mathcal{S}$}{
    \ForEach{frame $I_t$ in $s_k$}{
        Detect faces using MTCNN/Dlib\;
    }
    \If{at least one face is detected in $s_k$}{
        $\mathcal{S}_f \leftarrow \mathcal{S}_f \cup \{s_k\}$\;
    }
}

\tcp{Landmark extraction and graph-based temporal modeling}

$\mathbf{G}_{seq} \leftarrow [\,]$ \tcp*{shot-level GAT embeddings}

\ForEach{shot $s_k \in \mathcal{S}_f$}{
    $\mathcal{Z}_k \leftarrow [\,]$ \tcp*{frame-level GAT features in shot $k$}
    
    \ForEach{frame $I_t$ in $s_k$ with detected face}{
        $\mathbf{L}_t \leftarrow \text{SPIGA}(I_t)$ \tcp*{68-point landmarks}
        Construct graph $G_t = (\mathbf{L}_t, E)$ using anatomical edges\;
        $\mathbf{z}_t \leftarrow \text{GAT}(G_t)$ \tcp*{facial graph embedding}
        Append $\mathbf{z}_t$ to $\mathcal{Z}_k$\;
    }
    
    \If{$\mathcal{Z}_k$ is not empty}{
        $\mathbf{g}_k \leftarrow \text{MeanPool}(\mathcal{Z}_k)$ \tcp*{shot-level landmark feature}
        Append $\mathbf{g}_k$ to $\mathbf{G}_{seq}$\;
    }
}

$\mathbf{h}_{LSTM} \leftarrow \text{LSTM}(\mathbf{G}_{seq})$\;
$\mathbf{F}_{land} \leftarrow \text{GRU}(\mathbf{h}_{LSTM})$ \tcp*{temporal landmark embedding}

\tcp{Spatiotemporal visual feature extraction}

$\mathcal{V}_{seq} \leftarrow [\,]$\;

\ForEach{shot $s_k \in \mathcal{S}_f$}{
    Sample/resize $s_k$ into clip $C_k$\;
    $\mathbf{v}_k \leftarrow \text{3D-ResNet18}(C_k)$\;
    $\mathbf{v}_k \leftarrow \text{AdaptiveAvgPool}(\mathbf{v}_k)$\;
    Append $\mathbf{v}_k$ to $\mathcal{V}_{seq}$\;
}

$\mathbf{F}_{vis} \leftarrow \text{MeanPool}(\mathcal{V}_{seq})$ \tcp*{visual embedding}

\tcp{Dynamic weighted fusion of modalities}

$\alpha \leftarrow \sigma(\text{FC}([\mathbf{F}_{vis} \Vert \mathbf{F}_{land}]))$ \tcp*{$\alpha \in [0,1]$}
$\mathbf{F}_{fused} \leftarrow \alpha\,\mathbf{F}_{vis} + (1-\alpha)\,\mathbf{F}_{land}$\;

$\hat{y} \leftarrow \text{Classifier}(\mathbf{F}_{fused})$ \tcp*{final prediction}

\Return $\hat{y}$
\end{algorithm}

\begin{figure*}[t]
    \centering
    \includegraphics[width=\textwidth]{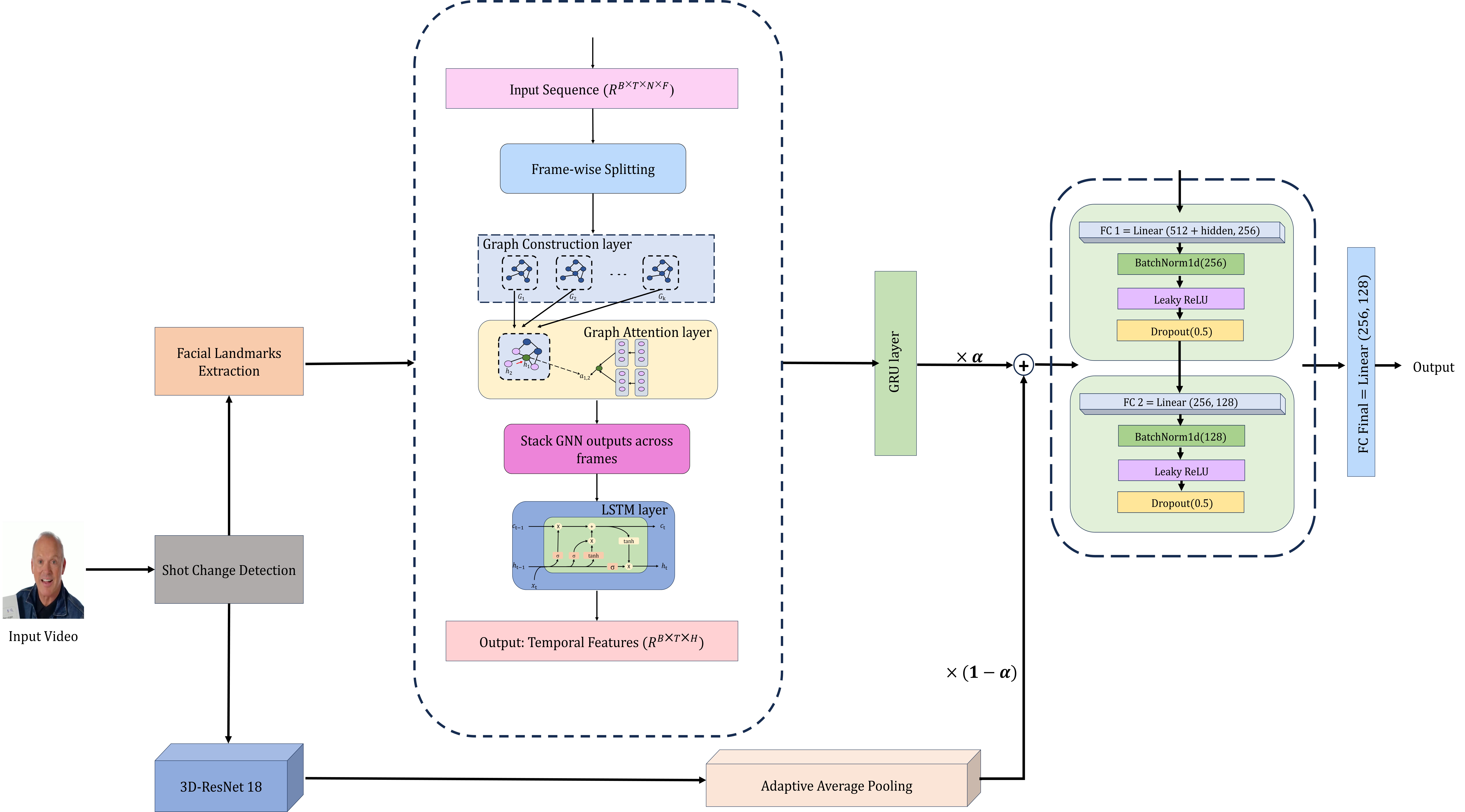}
    \caption{Architecture of the proposed model.}
    \label{fig:model_architecture}
\end{figure*}
\subsection{Shot Change Detection and Preprocessing}
The input to the proposed framework consists of a video sequence capturing an individual’s behavioral patterns. As an initial pre-processing step, shot change detection is performed using the TransNetV2 model \cite{soucek2024transnet}, which is designed to identify both abrupt and gradual transitions within video streams. As illustrated in Figure~2, TransNetV2 leverages Deep Dilated Convolutional Neural Network (DDCNN) cells with variable dilation rates to effectively capture multiscale spatiotemporal dependencies. In addition, a similarity network enhances transition detection by computing cosine similarities between frame pairs, incorporating both handcrafted features (e.g., RGB histograms) and learned feature representations. The model further employs multiple classification heads to improve the localization accuracy of transition points, thereby enhancing Shot boundary detection plays a pivotal role in structuring the video into semantically meaningful and temporally coherent segments, which are crucial for subsequent behavior analysis. To focus analysis on relevant content, only segments containing detectable faces are retained. Face detection is achieved through a hybrid approach that combines the strengths of MTCNN\cite{zhang2016mtcnn} and Dlib, enabling robust performance under challenging conditions such as varying illumination, occlusions, and head pose variations. In certain cases, either MTCNN or Dlib alone failed to detect faces due to extreme pose variations, partial occlusions, or other challenging conditions. To address this limitation, the two methods were employed in a complementary manner, such that a face was considered successfully detected if it was identified by either MTCNN or Dlib.

\subsection{Facial Landmark Extraction and Graph-Based Analysis}
To enable fine-grained analysis of facial dynamics, this study employs facial landmark extraction followed by graph-based processing. Each retained video shot is processed using the SPIGA model \cite{prados2022shape} to extract 68 facial landmarks. As shown in Figure~3, SPIGA is a modular architecture designed for robust and precise detection, featuring a Multi-Task Network (MTN) backbone with four stacked hourglass encoder-decoder modules that jointly estimate head pose and generate initial landmark predictions via a projected 3D face mesh. The model outputs heatmaps and visual features optimized using smooth L1 and Awing losses. Landmark refinement is performed through a cascade of Graph Attention Network (GAT)-based regressors, which treat landmarks as graph nodes and iteratively enhance localization by integrating visual and positional features through dynamic attention. A progressive reduction in the feature window size allows increasingly fine-grained detection. While SPIGA supports up to 98 landmarks, the 68-point configuration (Fig.~4) is adopted in this work due to hardware and resource limitations, ensuring feasible deployment within the available computational environment and real-time video analysis requirements.
\begin{figure*}[t]
    \centering
    \includegraphics[height=0.25\textheight, keepaspectratio]{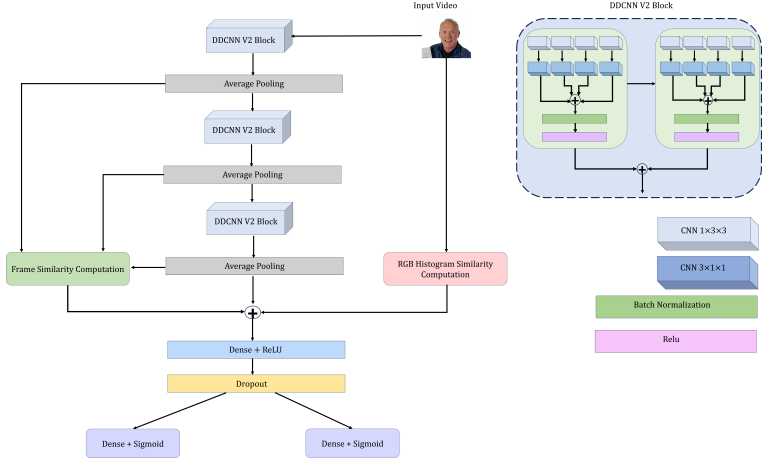}
    \caption{Architecture of the TransNetV2 model.}
    \label{fig:transnet_arch}
\end{figure*}

Following facial landmark extraction, each frame is represented as an undirected graph, where nodes correspond to individual facial landmarks and edges encode predefined spatial relationships that reflect the underlying anatomical structure of the human face. This graph-based formulation not only preserves the static geometry of the face but also enables the modeling of dynamic deformations over time as the facial configuration changes with expressions or behavior. To extract meaningful spatial features from these graph representations, a Graph Attention Network (GAT) is employed. In contrast to traditional Graph Convolutional Networks (GCNs), which apply uniform weighting to a node’s neighbors, GATs introduce a learnable attention mechanism that assigns varying levels of importance to each neighboring node during the aggregation process\cite{velivckovic2017graph}. This allows the network to selectively emphasize semantically rich and behaviorally relevant facial regions, such as the eyes, mouth, and brow, which are particularly sensitive to subtle cues associated with alcohol influence.

For each video frame, the GAT processes the corresponding landmark graph and produces a feature representation that captures both local topology and context-aware spatial relationships. These frame-level outputs are then temporally aligned and aggregated across the video segment. To construct a unified spatial descriptor for the entire segment, frame-wise averaging is applied across the GAT outputs, ensuring temporal consistency while mitigating frame-level noise. This spatial aggregation yields a compact yet expressive feature vector that encapsulates the overall facial configuration within a shot.
To capture temporal dynamics across the sequence of segments, the aggregated spatial features are fed into a sequential modeling pipeline consisting of a Long Short-Term Memory (LSTM) layer followed by a Gated Recurrent Unit (GRU) layer. The LSTM component is responsible for modeling long-range dependencies and preserving temporal context, effectively capturing evolving patterns in facial behavior. The subsequent GRU layer refines this representation by focusing on shorter-term dynamics and mitigating redundancy in the hidden state evolution. This two-stage temporal aggregation framework allows the model to learn both coarse and fine-grained temporal cues, enhancing its ability to distinguish between subtle behavioral shifts over time that may be indicative of alcohol impairment.

\begin{figure*}[htbp]
    \centering
    \includegraphics[height=5cm]{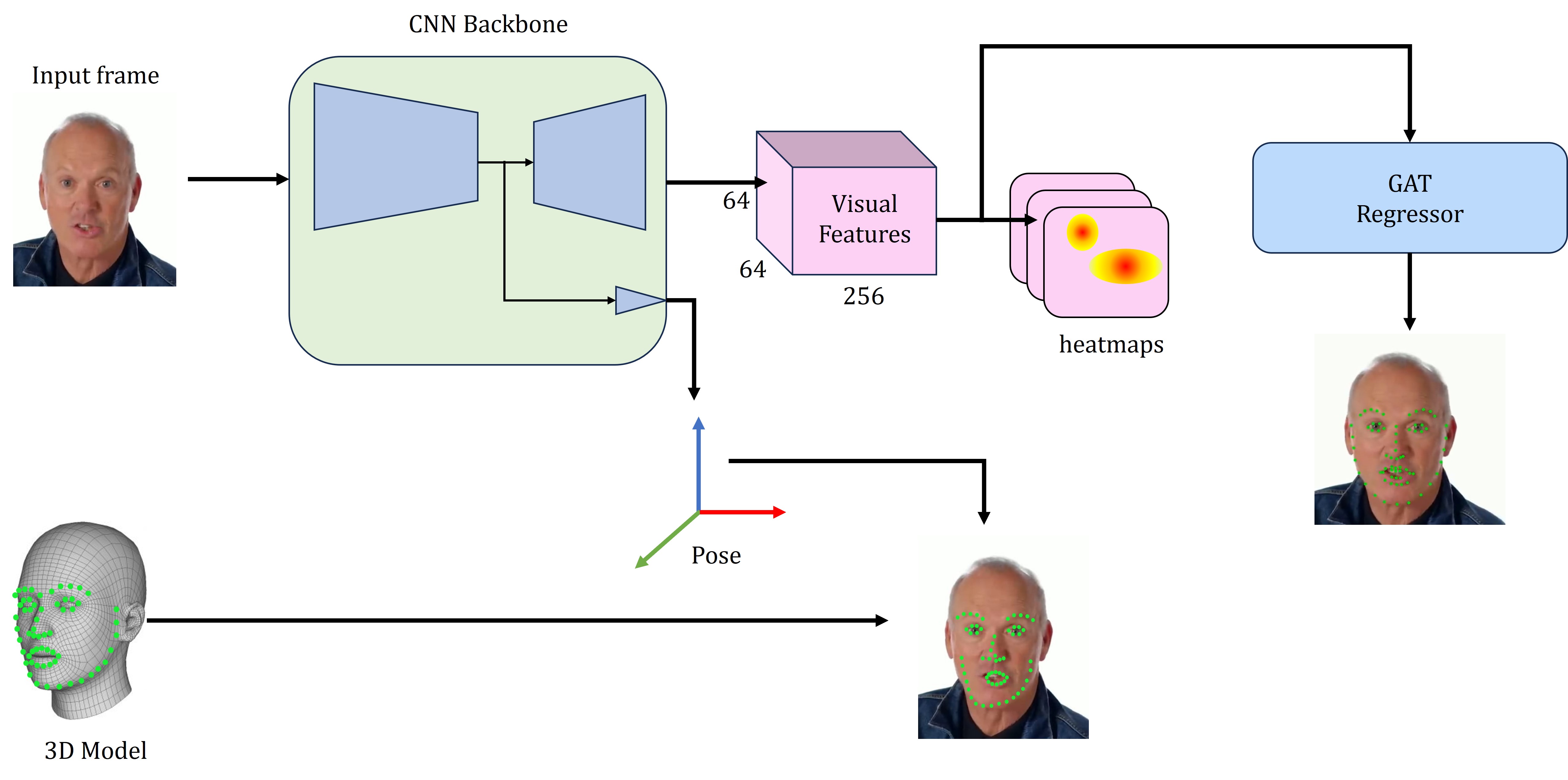}
    \caption{Architecture of SPIGA.}
    \label{fig:spiga_architecture}
\end{figure*}

\begin{figure}[t]
    \centering
    \includegraphics[width=0.25\linewidth]{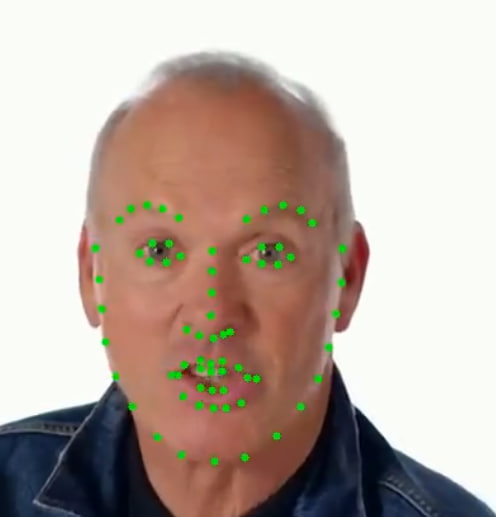}
    \caption{The 68 facial landmark points and their positioning }
    \label{fig:enter-label}
\end{figure}
\subsection{Spatiotemporal Visual Feature Extraction}
In parallel with the graph-based analysis of facial landmarks, the proposed framework incorporates a complementary visual stream to extract holistic spatiotemporal features directly from the original video segments. This stream is powered by a 3D ResNet-18 architecture\cite{tran2018spatiotemporal}, which extends the traditional 2D ResNet by replacing standard 2D convolutional kernels within its residual blocks with 3D convolutions. This architectural modification enables the network to jointly model spatial and temporal dimensions, allowing it to capture motion patterns, temporal consistency, and inter-frame dependencies that are essential for understanding dynamic behaviors.
The 3D ResNet architecture retains the core residual learning mechanism of its 2D counterpart, wherein identity shortcut connections are used to bypass one or more convolutional layers. As a result, the model is able to learn richer hierarchical representations that encode not only static appearance features, such as facial texture and background cues, but also dynamic patterns related to body posture, motion blur, head sway, and gaze drift, which are particularly relevant in the context of intoxication assessment.
To enhance the model’s ability to generalize across diverse subjects and video conditions, the 3D ResNet is initially pretrained on the large-scale Kinetics action recognition dataset\cite{kay2017kinetics}. This pretraining allows the network to acquire domain-agnostic motion and appearance priors that are transferable to the alcohol detection task\cite{Tran2018CVPR}. During fine-tuning, the model is adapted to the specific dataset and objective of intoxication classification. This transfer learning strategy accelerates convergence and mitigates overfitting, which is particularly critical when training on domain-specific datasets of limited scale.

\subsection{Adaptive Feature Fusion}
A key innovation of the proposed framework is centered on its adaptive weighted fusion mechanism, which integrates multimodal feature representations derived from two complementary branches: (i) temporal embeddings extracted from facial landmark sequences via the GAT+LSTM pipeline, and (ii) spatiotemporal features obtained from the 3D ResNet architecture. This fusion strategy is designed to dynamically balance the contributions of the two modalities, ensuring robustness under varying input conditions. This fusion is governed by a learnable scalar $\alpha$, obtained via a sigmoid activation to ensure that $\alpha \in [0,1]$ , which modulates the relative weight assigned to each modality. Formally, the final fused feature representation is computed as:
\begin{equation}
F_{fused} = \alpha F_{visual} + \left(1 - \alpha\right) F_{landmark}
\end{equation}
Where \( F_{visual}\) and  \( F_{landmark}\) denote the feature vectors produced by the 3D ResNet and the GAT+LSTM pipeline, respectively. This formulation allows the network to learn an optimal weighting during training, adapting to the reliability and discriminative power of each modality.
Such a mechanism proves especially beneficial in real-world scenarios where input quality may be inconsistent. For example, in cases of poor illumination, motion blur, or background clutter conditions under which appearance-based features from the 3D ResNet may degrade the model can increase reliance on the structural and temporal cues encoded in the facial landmark stream. Conversely, when landmark detection is impaired due to occlusion, extreme head poses, or partial visibility, the visual modality can compensate by leveraging residual texture and motion patterns. This adaptive fusion not only enhances robustness to noise and occlusion but also improves the model’s ability to generalize across diverse environmental and subject-level variations.

\section{\textbf{Experimental Results} }
\subsection{Dataset}

A new video dataset was curated from YouTube to support the development and evaluation of the proposed multimodal intoxication-detection model. This was required because many source videos from the DIF dataset [7] were no longer accessible, preventing reproducibility. A total of 202 raw videos (101 sober, 101 intoxicated) were collected and segmented using the TransNetV2 shot-boundary detector, producing 3,542 shot-level clips. To mitigate class imbalance introduced by segmentation, an equal number of samples was selected per class, yielding 1,771 clips each. Representative samples are shown in Fig.~5.

The dataset was constructed with attention to demographic diversity. The gender distribution is 56\% women and 44\% men, with subjects spanning a broad range of apparent ages. Racial diversity includes six major groups: Caucasian, Asian, Middle Eastern, Mexican, African American, and Indian. All videos were standardized to 720×1280 resolution to ensure consistent visual quality for multimodal feature extraction. Table~1 summarizes the dataset characteristics. 
\begin{table}[htbp]
\centering
\caption{Dataset Characteristics}
\label{tab:dataset_characteristics}
\vspace{10pt}
\begin{tabular}{ll}
\hline
\textbf{Characteristic}    & \textbf{Description}  \\ \hline \\[2pt]
Total Video Segments       & 3,542 \\
                           &      \\
Class Distribution         & \begin{tabular}[t]{@{}l@{}}Class 1 (Intoxicated): 1,771 \\ Class 2 (Sober): 1,771\end{tabular} \\
                           &      \\
Gender Distribution        & \begin{tabular}[t]{@{}l@{}}Women: 56\% \\ Men: 44\%\end{tabular} \\
                           &      \\
Age Groups                 & \begin{tabular}[t]{@{}l@{}}Under 21 years \\ 21--40 years \\ Over 40 years\end{tabular} \\
                           &      \\
Racial Groups              & \begin{tabular}[t]{@{}l@{}}Caucasian, Asian, Middle Eastern, \\ Mexican, African American, Indian\end{tabular} \\
                           &      \\
Video Resolution           & 720 $\times$ 1280 pixels \\
                           &      \\
Dataset Split              & \begin{tabular}[t]{@{}l@{}}Training: 80\% \\ Testing: 20\% \\ Subject-independent split to prevent leakage\end{tabular} \\ \hline
\end{tabular}
\end{table}
\begin{figure}[t]
    \centering
    \includegraphics[width=0.5\linewidth]{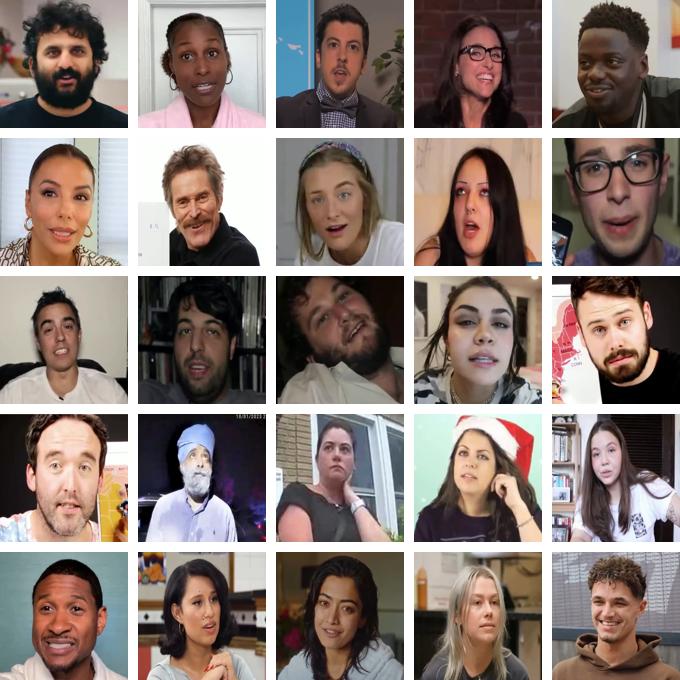}
    \caption{Sample frames from our intoxication detection dataset }
    \label{fig:enter-label}
\end{figure}
\subsection{Implementation Details }
Following adaptive fusion, the multimodal feature vector (512-dim) was compressed through a two-stage reduction pipeline (512→256→128) with batch normalization, dropout (0.5), and Leaky ReLU activation. The final 128-dim representation was mapped to a binary output (intoxicated vs. sober) using a fully connected layer with cross-entropy loss. Model training was performed with the Adam optimizer (learning rate = 1e-4, weight decay = 1e-5) over 25 epochs. All experiments were implemented in PyTorch and executed on an NVIDIA GeForce GTX TITAN X GPU.

\subsection{Evaluation}
In this section, we evaluate the proposed framework against baseline methods. The dataset was divided using an 80/20 subject-independent split to prevent identity-based leakage, ensuring fair comparison. Standard classification metrics (accuracy, precision, recall) are employed, alongside training dynamics and confusion matrix analysis, to provide both quantitative and qualitative insights into model performance. 
As shown in Table~2 ,the performance of the proposed model was benchmarked against two baseline approaches: a customized 3D Convolutional Neural Network (3D-CNN) and a VGGFace+LSTM architecture\cite{mehta2019dif}. As shown in Table~7, the proposed model outperformed both baselines, achieving an accuracy of 95.82\%, compared to 87.7\% for the 3D-CNN and 56.0\% for the VGGFace+LSTM. Model performance was assessed using standard classification metrics, defined as:

\begin{equation}
\text{Accuracy} = \frac{TP + TN}{TP + TN + FP + FN}
\end{equation}

\begin{equation}
\text{Precision} = \frac{TP}{TP + FP}
\end{equation}

\begin{equation}
\text{Recall} = \frac{TP}{TP + FN}
\end{equation}
where TP, TN, FP and FN denote the number of true positives, true negatives, false positives, and false negatives, respectively. 
In addition to overall accuracy, the proposed model achieved superior precision and recall scores (both 0.977), demonstrating robustness in correctly identifying intoxicated individuals while minimizing misclassifications. Training and validation dynamics are shown in Fig.~6, while the confusion matrix in Fig.~7 provides class-wise insights. Most errors occurred in ambiguous cases, such as sober individuals with prolonged eye closure or atypical head poses, underscoring the complexity of real-world scenarios.
\begin{figure}[htbp]
    \centering
    \includegraphics[width=0.85\linewidth, height=4cm
]{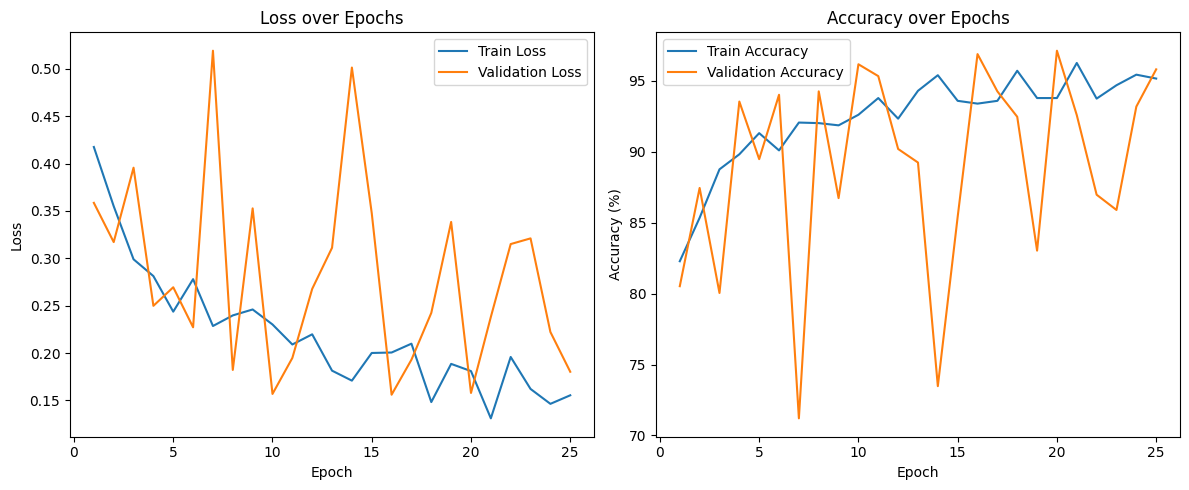}
    \caption{Training and Validation Accuracy and Loss Curves of the Proposed Model }
    \label{fig:enter-label}
\end{figure}
\begin{figure}[htbp]
    \centering
    \includegraphics[width=0.6\linewidth]{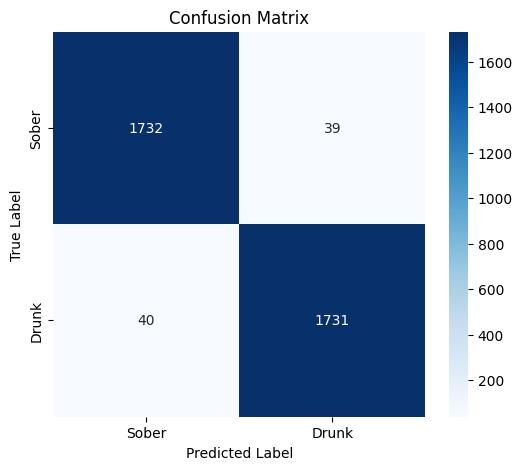}
    \caption{Confusion matrix of the  Proposed Model }
    \label{fig:enter-label}
\end{figure}
The strong performance of the proposed model stems from its hybrid architecture, which integrates spatiotemporal features from a 3D ResNet with temporal facial geometry from landmark tracking. Unlike the VGGFace+LSTM, which emphasizes static appearance, our model captures both macro-level dynamics (e.g., posture changes) and micro-level cues (e.g., eye and mouth movements). This multi-scale representation enables broader generalization to diverse behavioral indicators of intoxication. 
To further enhance robustness, a dynamic weighted fusion mechanism was employed. A learnable fusion parameter ($\alpha = 0.4704$) allowed the system to prioritize the most informative modality depending on input quality. This adaptivity proved particularly effective under challenging conditions such as occlusion, pose variation, and lighting inconsistencies.

\begin{table}[!t]
    \centering
    \caption{Performance Comparison of the Proposed Model and Previous Methods}
    \vspace{5pt}
    \footnotesize
    \begin{tabular}{lccc}
        \toprule
        \textbf{Model} & \textbf{Accuracy} & \textbf{Precision} & \textbf{Recall} \\
        \midrule
        Customized 3D-CNN \cite{mehta2019dif} & 86.7\% & 0.88 & 0.88 \\
        VGGFace + LSTM \cite{mehta2019dif}    & 56.0\% & 0.79 & 0.50 \\
        \textbf{Proposed Model}               & \textbf{95.82\%} & \textbf{0.97} & \textbf{0.97} \\
        \bottomrule
    \end{tabular}
    \label{tab:perf_compare}
\end{table}

In contrast, prior models faced notable limitations: the 3D-CNN incurred high computational costs and required large-scale data for generalization, while VGGFace+LSTM lacked sensitivity to temporal variation. By comparison, the proposed framework integrates spatial, temporal, and contextual cues in a unified design, achieving state-of-the-art performance and demonstrating applicability to real-world intoxication detection.

\subsection{Ablation Studies}
A series of ablation studies was conducted to systematically evaluate the contribution of individual components within the proposed architecture. These experiments isolate the effects of key design elements including modality integration, activation functions, the weighted fusion mechanism, and the incorporation of auxiliary features on overall model performance. By selectively removing or altering each component, the relative importance and functional role of individual elements were quantitatively assessed. Performance was compared across configurations using standard evaluation metrics such as accuracy, precision, and recall, providing a rigorous basis for analysis. Details of these ablation studies are presented in the subsequent subsections.

\subsubsection{Effect of Combining Visual Features and Facial Landmarks}
The effectiveness of integrating visual features (extracted via 3D ResNet) with facial landmarks (processed through a Graph Attention Network and LSTM) was evaluated through a comparative analysis against unimodal baselines. As presented in Table~3, the proposed multimodal configuration achieved an accuracy of 95.82\%, surpassing both the landmark-only model (95.00\%) and the 3D ResNet-only model (94.86\%). These results underscore the complementary nature of the two modalities: facial landmarks provide fine-grained structural and motion-related information, while visual features offer richer spatial context and texture-level cues. The proposed weighted fusion mechanism effectively leverages these complementary strengths, resulting in improved overall classification performance.
\begin{table}[htbp]
    \centering
     \caption{Effect of Modality Integration
}\vspace{5pt}\begin{tabular}{cccc}\toprule
         \textbf{Model}&  \textbf{Accuracy}&  \textbf{Precision}& \textbf{Recall}\\\midrule
         3D-ResNet&  87.7\%&  0.88& 0.88\\
         Landmarks only&  95.00\%&  0.95& 0.95
\\
        \textbf{ Proposed Combined(fusion)}&  \textbf{95.82\%}&  \textbf{0.977}& \textbf{0.977}\\ \bottomrule
    \end{tabular}
   
    \label{tab:my_label}
\end{table}
\subsubsection{Effect of Activation Functions}
To assess the influence of activation functions on learning dynamics and model performance, three widely used functions ReLU, Swish, and Leaky ReLU were evaluated within the combined model architecture. As reported in Table~4, the Leaky ReLU activation function yielded the highest classification accuracy at 93.7\%, outperforming both ReLU (93.05\%) and Swish (93.24\%). The superior performance of Leaky ReLU can be attributed to its ability to mitigate the “dying ReLU” problem by preserving a small, non-zero gradient for negative input values. This property promotes more stable gradient flow during training, leading to improved convergence behavior and overall model robustness. 
\begin{table}[htbp]
\centering
\footnotesize
\caption{Effect of Activation Function}
\begin{tabular}{cc}
\toprule
Model & Accuracy \\
\midrule
Fused Model + ReLU      & 93.05\% \\
Fused Model + Swish     & 93.24\% \\
\textbf{Fused Model + LeakyReLU}& \textbf{93.70\%}\\
\bottomrule
\end{tabular}
\label{tab:activation}
\end{table}

\subsubsection{Effect of Weighted Fusion}
A learnable weighted fusion mechanism was introduced to dynamically balance the contributions of the visual and facial landmark streams. This adaptive approach was compared against a baseline method involving simple feature concatenation. As shown in Table~5, the weighted fusion strategy yielded a substantial performance gain, increasing classification accuracy from 93.7\% (with Leaky ReLU and concatenation) to 95.82\% when combined with the proposed fusion mechanism. These results underscore the critical role of adaptive feature integration in multimodal learning, allowing the model to selectively emphasize the most informative modality under varying input conditions.

 \begin{table}[htbp]
     \centering
        \caption{Effect of Weighted Fusion
}\vspace{5pt}\begin{tabular}{cc}\toprule
          Model& Accuracy
\\\midrule
          Fused Model + LeakyReLU     & 93.70\%\\
          \textbf{Weighted Fusion + LeakyReLU}& \textbf{95.825}\\ \bottomrule
     \end{tabular}

     \label{tab:my_label}
 \end{table}
\subsubsection{ Effect of Eye and Mouth Aspect Ratios (EAR , MAR)}
To further investigate the utility of Eye Aspect Ratio (EAR) as a discriminative feature for intoxication detection\cite{tran2018spatiotemporal}, we visualized its distribution across the sober and intoxicated classes, as shown in Fig.~8. The box plot (top) and scatter plot (down) reveal substantial overlap in EAR values between the two conditions, with no distinct separation or class-specific trends. This empirical evidence reinforces the quantitative findings reported in Table~6, where the inclusion of EAR yielded minimal improvements in model performance. 
These results suggest that EAR, while capturing certain geometric properties of the eye region, does not offer discriminative power beyond what is already encoded in the graph-based facial representation. Consequently, its integration appears redundant in this context and underscores the necessity of feature selection strategies that prioritize complementary and non-redundant cues for behavioral recognition tasks. 
\begin{figure}[t]
    \centering
    \includegraphics[width=0.95\columnwidth, height=4cm, keepaspectratio]{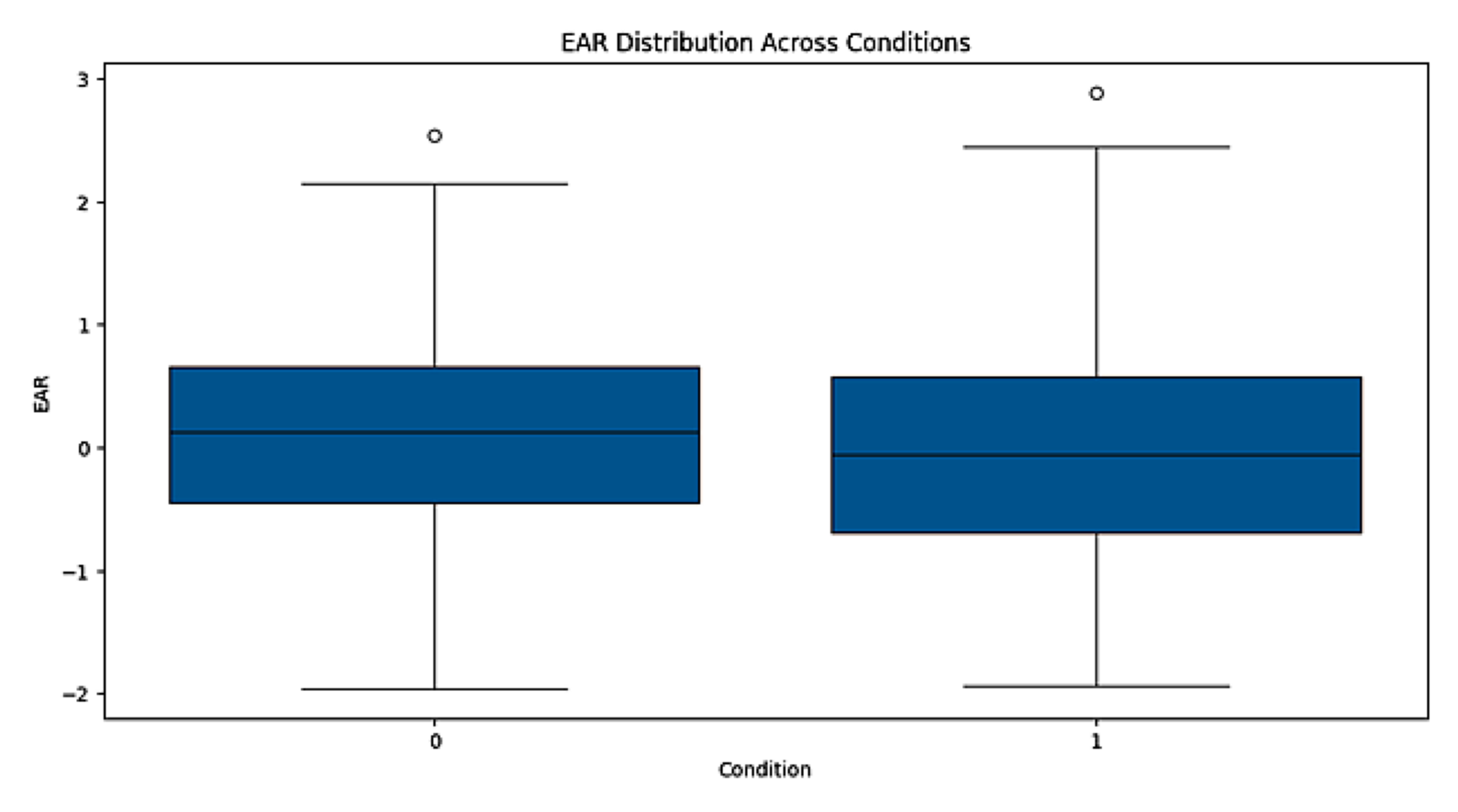}
    \includegraphics[width=0.95\columnwidth, keepaspectratio]{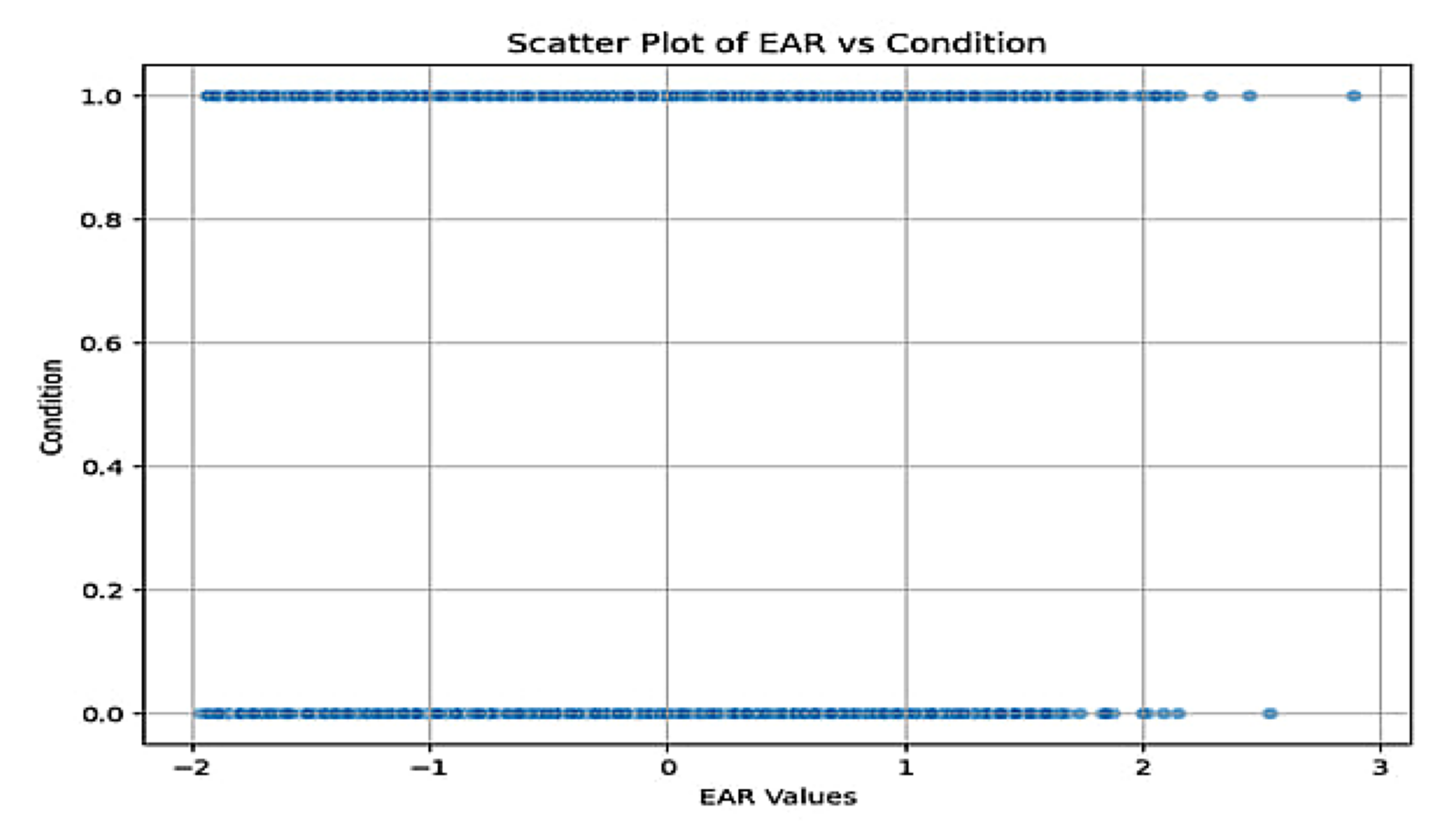}
    \caption{Distribution of EAR across sober and intoxicated conditions.}
    \label{fig:minipage_images}
\end{figure}
\begin{table}[htbp]
    \centering
     \caption{Effect of Eye and Mouth Aspect Ratios
}\vspace{5pt}\begin{tabular}{cccc}\toprule
         Model&  Accuracy&  Precision& Recall
\\\midrule
         \textbf{Landmarks only}&  \textbf{95.00\%}&  \textbf{0.95}& \textbf{0.95}\\
         Landmarks + EAR&  94.00\%&  0.94& 0.95
\\
         Landmarks + MAR&  85.00\%&  0.87& 0.85
\\ \bottomrule
    \end{tabular}
   
    \label{tab:my_label}
\end{table}
\subsubsection{Effect of Demographic Features}
To investigate potential demographic influences on model performance, auxiliary attributes; specifically age (categorized as above or below 35), gender, and race (across six predefined groups), were extracted using the DeepFace framework \cite{serengil2021lightface} and integrated into the classification pipeline. However, as shown in Table 7, the inclusion of these features led to a modest reduction in overall accuracy, declining from 95.82\% to 94.00\%.
This decline can be explained by the static nature of DeepFace’s attribute estimation, which lacks temporal sensitivity and does not align well with the dynamic behavioral patterns essential for video-based recognition.  Although demographic attributes are known to influence physiological responses to alcohol, for example, higher blood alcohol levels and prolonged effects in women, increased sensitivity in older adults, and race-specific reactions such as alcohol-induced flushing in Southeast Asians \cite{aw1986racial},  these effects do not consistently manifest in observable facial behavior. Consequently, when modeled without temporal context, demographic features may introduce confounding variability rather than improve predictive performance. These findings highlight the importance of aligning auxiliary inputs with both the temporal dynamics of the task and the behavioral manifestations relevant to intoxication detection. 
\begin{table}[htbp]
    \centering
      \caption{Effect of Demographic Features } \vspace{5pt}\begin{tabular}{cccc}\toprule
         Model&  Accuracy&  Precision& Recall
\\\midrule
         \textbf{Landmarks only}&  \textbf{95.00\%}&  \textbf{0.95}& \textbf{0.95}\\
         Landmarks + Demographic Features&  94.00\%&  0.94& 0.94
\\ \bottomrule
    \end{tabular}
  
    \label{tab:my_label}
\end{table}

\subsection{Model interpretability}
To enhance the transparency and explainability of the proposed model, a series of explainability analyses were performed using Gradient-weighted Class Activation Mapping  (Grad-CAM)\cite{selvaraju2017gradcam} and salient point extraction.These techniques provide visual and structural insights into the model's decision-making process by identifying the regions or features that contributed most significantly to its predictions. 
\subsubsection{Grad-CAM Visualization}
To interpret the decision-making process of the proposed model, Grad-CAM was applied to the output of the 3D-ResNet component, enabling the generation of class-specific activation heatmaps across individual video frames. As illustrated in Figure~9, these visualizations consistently demonstrated that the model's attention was concentrated on key facial regions, particularly around the eyes, mouth, and jawline. These areas are clinically and behaviorally associated with signs of intoxication, such as prolonged eye closure, delayed blinking, and irregular mouth movements. The alignment between the model’s learned focus and known physiological indicators of alcohol consumption supports the interpretability and reliability of the model, reinforcing its suitability for real-world behavioral assessment applications. 
\begin{figure}[htbp]
    \centering
    \includegraphics[width=0.5\linewidth]{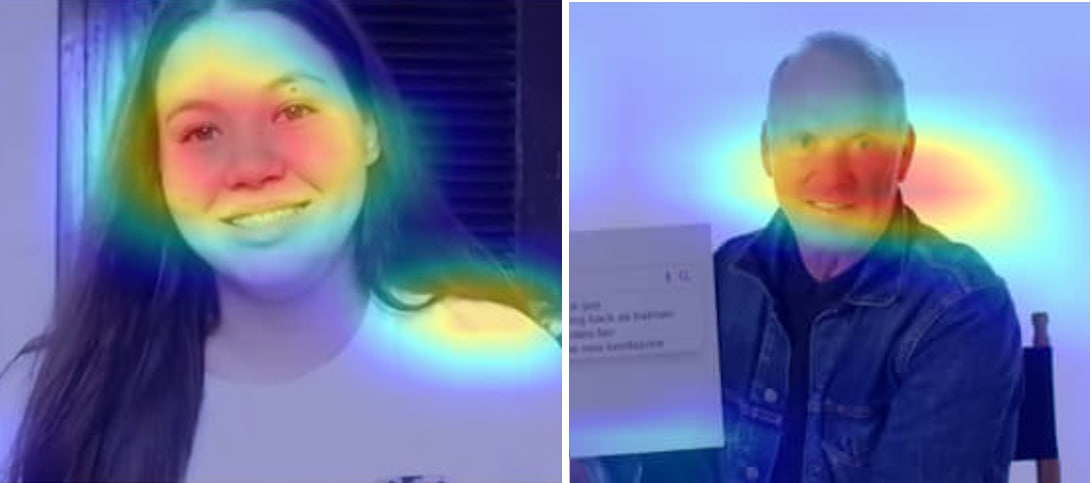}
    \caption{Model interpretability visualization }
    \label{fig:enter-label}
\end{figure}
\subsubsection{Salient Landmark Point Analysis}
A gradient-based sensitivity analysis was conducted on the facial landmark data to identify the most influential keypoints contributing to classification decisions. This analysis involved computing normalized importance scores via backpropagation, allowing for the quantification of each landmark's contribution to the model’s output. As summarized in Table~8, the highest sensitivity scores were observed in the jawline region (points 12–15), the eyes (points 43–46), and the mouth corners (points 49–54). These findings indicate that the model primarily relies on behaviorally relevant facial regions, those associated with head movements\cite{Modig2012Alcohol}, blinking patterns\cite{Makowski2023Alcohol}, and mouth movement\cite{Pisoni1989AlcoholSpeech}, which are known indicators of alcohol-induced impairment.
\begin{table}[htbp]
    \centering
      \caption{Most Influential Facial Landmark Points }\vspace{5pt}\begin{tabular}{cc}\toprule
         Key Facial Regions& Normalized Importance Score
\\\midrule
         Jawline (points 12–15)& 0.40 – 0.50
\\
         Eyes (points 43–46)
& 0.20 – 0.30\\
         Mouth corners (49–54)& 
0.15 – 0.18\\ \bottomrule
    \end{tabular}
  
    \label{tab:my_label}
\end{table}

\section{Discussion}

The proposed multi-modal framework demonstrates notable strengths while also revealing key areas for refinement. A major advantage lies in its two-stage temporal aggregation, which captures both coarse and fine-grained behavioral dynamics, thereby improving sensitivity to subtle cues of alcohol impairment. The adaptive fusion strategy further enhances robustness against noise and occlusion and enables consistent performance across diverse environments and subjects. Together, these mechanisms contributed to the model’s strong accuracy of 95.82\% and its superiority over baseline methods.

Nonetheless, misclassification analysis identified two primary error modes. False positives often arose when sober participants exhibited behaviors such as prolonged eye closure, downward gaze, or head tilting, movements resembling intoxication cues. Grad-CAM results showed that, in these cases, the model sometimes concentrated on less informative regions (e.g., forehead), reducing discriminative power. False negatives typically occurred among intoxicated individuals with neutral expressions or minimal facial movement, which limited the temporal signals required for reliable detection.

A further limitation emerged with demographic features. Incorporating attributes such as age, gender, and race reduced accuracy from 95.82\% to 94.00\%. This decline likely reflects the static nature of DeepFace-based estimation, which lacks temporal sensitivity and fails to align with dynamic behavioral patterns. While demographic factors are known to influence physiological responses to alcohol, these effects do not consistently manifest in observable facial behavior, introducing variability rather than predictive value when modeled statically.

In summary, the framework effectively integrates temporal modeling and adaptive fusion to deliver state-of-the-art performance in intoxication detection, yet remains sensitive to behavioral ambiguities and demographic variability. Future work should address these challenges by refining attention mechanisms, incorporating temporally contextualized auxiliary features, and exploring transformer-based architectures. Given its adaptability, the framework may also extend to related domains such as fatigue, drowsiness, and affective state recognition.

\section{\textbf{Conclusion and Future Works}
}
This study proposed a novel multi-modal framework for alcohol consumption detection in real-world scenarios by integrating spatiotemporal visual feature extraction with facial landmark-based temporal modeling. A demographically and environmentally diverse video dataset was curated to ensure robustness and generalization. The architecture dynamically fuses 3D-ResNet features with facial dynamics, achieving an accuracy of  95.82\%, with precision and recall both at 0.977, thereby surpassing baseline models. Extensive experiments and ablation studies confirmed the efficacy of combining spatial and temporal cues to capture subtle behavioral indicators of intoxication, offering a scalable, non-invasive alternative to traditional detection methods.
Future work will focus on enhancing the system’s behavioral interpretation and reliability by incorporating features such as eye movement, head pose, and interaction dynamics, alongside advanced spatiotemporal architectures like video-based transformers (e.g., Video Swin Transformer). Further efforts will expand dataset diversity and improve demographic attribute extraction (e.g., age, gender, race) to deepen contextual understanding and support more robust deployment in safety-critical applications. Moreover, with appropriate modifications, the proposed framework could be extended beyond intoxication detection to related domains such as fatigue monitoring and depression detection, where subtle facial and behavioral cues also play a crucial role.

\bibliographystyle{IEEEtran} 
\bibliography{references}
\end{document}